\documentclass[10pt, conference, compsocconf]{IEEEtran}
\ifCLASSINFOpdf
\else
\fi
\usepackage{amsmath}
\usepackage{algorithmic}
\usepackage{array}
\usepackage{url}
\usepackage{bm}
\usepackage{times}
\usepackage{epsfig}
\usepackage{graphicx}
\usepackage{float}
\floatstyle{plaintop}
\restylefloat{table}
\usepackage{tabularx}
\usepackage{amsmath}
\usepackage{amsfonts}

\usepackage{multirow}
\newcommand{\xmark}{\ding{55}}%
\newcommand{\cmark}{\ding{51}}%
\usepackage[bottom]{footmisc}
\usepackage{pifont}
\usepackage{makecell}
\def\HS{\hspace{\fontdimen2\font}}

\begin{document}
\title{Domain Generalization via Universal Non-volume Preserving Approach}

\author{Dat T. Truong$^{1, 3, 4}$, Chi Nhan Duong$^{2}$, Khoa Luu$^{1}$, Minh-Triet Tran$^{3, 4}$, Ngan Le$^{1}$\\
				$^{1}$ University of Arkansas, USA\\
				$^{2}$ Concordia University, Canada\\
			    $^{3}$ University of Science, Ho Chi Minh city, Vietnam\\
			    $^{4}$ Vietnam National University, Ho Chi Minh city, Vietnam\\			    
		        \small\texttt{\{tt032, khoaluu, thile\}@uark.edu, dcnhan@ieee.org, tmtriet@fit.hcmus.edu.vn}}


\maketitle

\begin{abstract}
Recognition across domains has recently become an active topic in the research community. However, it has been largely overlooked in the problem of recognition in new unseen domains. Under this condition, the delivered deep network models are unable to be updated, adapted, or fine-tuned. Therefore, recent deep learning techniques, such as domain adaptation, feature transferring, and fine-tuning, cannot be applied. This paper presents a novel approach to the problem of domain generalization in the context of deep learning.
The proposed method\footnote{Source code will be publicly available.} is evaluated on different datasets in various problems, i.e. (i) digit recognition on MNIST, SVHN, and MNIST-M, (ii) face recognition on Extended Yale-B, CMU-PIE and CMU-MPIE, and (iii) pedestrian recognition on RGB and Thermal image datasets. The experimental results show that our proposed method consistently improves performance accuracy. It can also be easily incorporated with any other CNN frameworks within an end-to-end deep network design for object detection and recognition problems to improve their performance.

\end{abstract}

% \begin{IEEEkeywords}
% component; formatting; style; styling;

% \end{IEEEkeywords}

% \fi
\IEEEpeerreviewmaketitle

\section{Introduction} \label{sec:intro}

\begin{figure}[t]
	\centering \includegraphics[width=0.99\columnwidth]{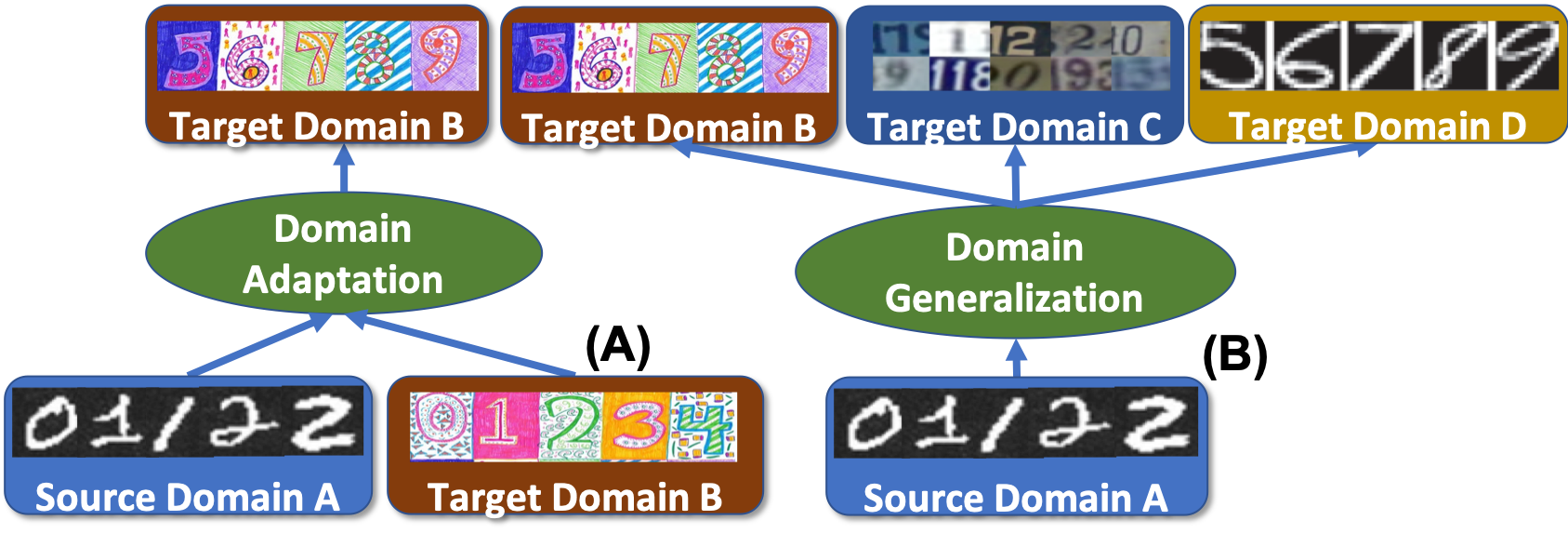}
	\caption{Comparison between domain adaptation (A) and our proposed domain generalization (B) problems}	
	\label{fig:Fig1}
\vspace{-5mm}
\end{figure}

\begin{figure}[t]
	\centering \includegraphics[width=0.99\columnwidth]{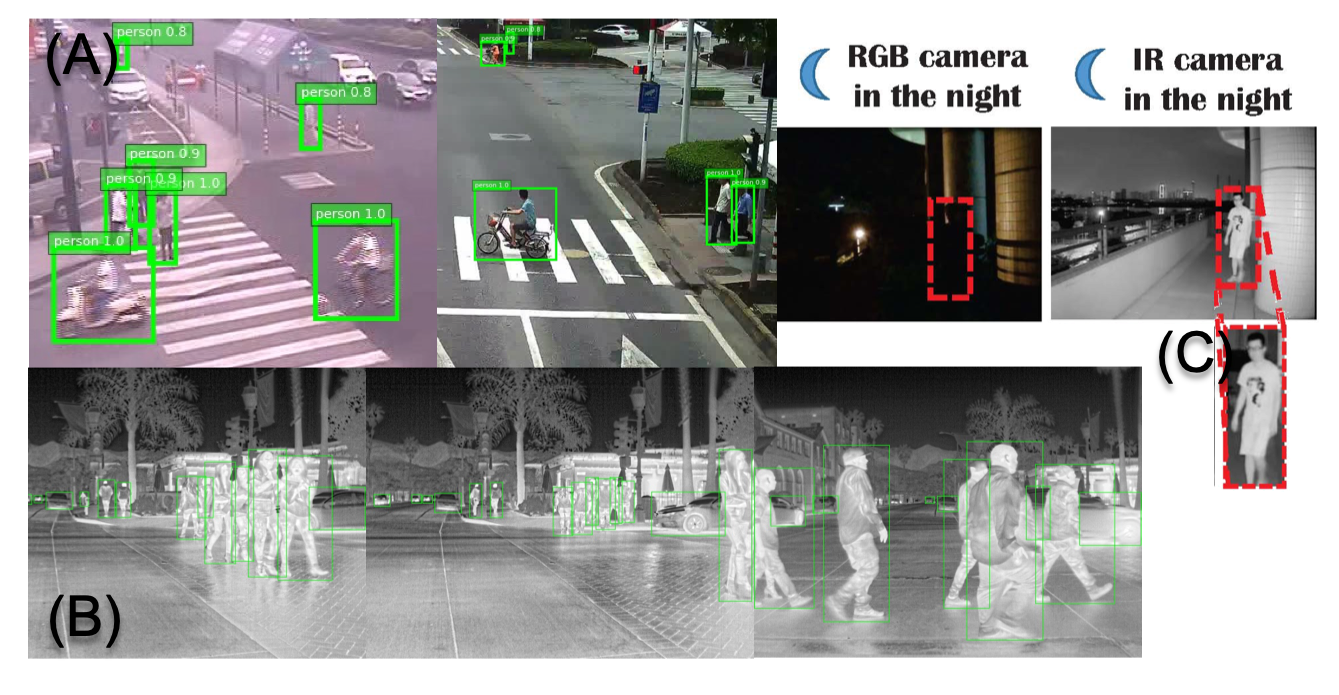}
	\caption{The ideas of domain generalization. The deep model is trained only in a single domain (A), i.e. RGB images. It is deployed in other unseen domains, i.e. thermal images (B) and infrared images (C).}	
	\label{fig:test_domains}
\end{figure}

Deep learning-based detection and recognition studies have been recently achieving very accurate performance in visual applications. However, many such methods assume the testing images come from the same distribution as the training ones and often fail when performing in new unseen domains. %For example, in face recognition, a system is trained on RGB images/videos and then deployed on infrared or thermal images/videos. 
Indeed, detection and classification crossing domains have recently become active topics in the research communities. In particular, \textit{domain adaptation} \cite{pmlr-v37-ganin15} \cite{adda_cvpr2017} has received significant attention in computer vision. 
In the domain adaptation (Fig. \ref{fig:Fig1}(A)), we usually have a large-scale training set with labels, i.e., the source domain A, and a small training set with or without labels, i.e., the target domain B. 
The knowledge from the source domain A is learned and adapted to the target domain B. During the testing time, the trained model is deployed \textit{only} in the target domain B. 
Recent results in domain adaptation have shown significant improvement in the many computer vision applications. 
However, the trained models are potentially deployed not only in the target domain B but also in many other \textit{new unseen} domains, e.g., C, D, etc. (Fig. \ref{fig:Fig1}(B)) in real-world applications. In these scenarios, the released deep network models are usually unable to be retrained or fine-tuned with the inputs in new unseen domains or environments, as illustrated in Fig. \ref{fig:test_domains}. 
Thus, domain adaptation cannot be applied in these problems since the new unseen target domains are unavailable. 
%In the context of the proposed problem, there is no available information about new unseen environments or domains where the system will be deployed.
%during training. % Moreover, domain adaptation methods only allow a pair of domains, i.e. the source domain and the target domain. Meanwhile,  real-world applications usually require more than just a pair of domains. In practice, the number of domains that released models are potentially deployed is usually large and unpredictable.

%{\color{blue} Should write like this?.In order to deal with the problem of unseen domains, \cite{triplet_loss, range_loss} paid attention to propose new loss functions to deal with hard samples considered as unseen domains. Instead of working on loss function, \cite{Huang_2017_CVPR} increased deep network structures to mine hard samples in training sets. }
  
Besides, there are some prior works to perform recognition problems with high accuracy by presenting new loss functions \cite{triplet_loss} \cite{range_loss} or increasing deep network structures \cite{deep_resnet} via mining hard samples in training sets. These loss functions are deployed to deal with hard samples considered as unseen domains.
%, one can consider it as hard-sample problems. %and then can be solved using new loss functions, e.g. Center Loss \cite{center_loss}, Range Loss \cite{range_loss}, etc. 
However, these methods are limited to be generalized in new unseen domains in real-world applications.
Some real-world problems are unable to observe training samples from new unseen domains in the training process.
Therefore, in the scope of this work, there is no assumption about the new unseen domains. 
Our proposed method can be supportively incorporated with Convolutional Neural Networks (CNNs)-based detection and classification methods to train within an end-to-end deep learning framework to improve the performance potential.

\subsection{Contributions of this Work}

This work presents a novel domain generalization approach to learn to better generalize new unseen domains. The restrictive setting is considered in this work where there is \textit{only single source domain available for training}. Table \ref{tab:TenMethodSumm} summarizes the differences between our approach and the prior works.
Our contributions can be summarized as follows.

A novel approach named \textit{Universal Non-volume Preserving (UNVP)} and its extension named \textit{Extended Universal Non-volume Preserving (E-UNVP)} frameworks are firstly introduced to generalize environments of new unseen domains from a given single-source training domain.
Secondly, the environmental features extracted from the environment modeling via Deep Generative Flows (DGF) and the discriminative features extracted from the deep network classifiers are then unified together to provide final generalized deep features that are robustly discriminative in new unseen domains. Our approach is designed within an end-to-end deep learning framework and inherits the power of the CNNs.
It can be quickly end-to-end integrated with a CNN-based deep network design for object detection or recognition to improve the performance.
Finally, the proposed method has experimented in various visual modalities and applications with consistently improving performances.

\section{Related Work} \label{sec:related_work}

\textbf{Domain Adaptation} has recently become one of the most popular research topics in the field \cite{pmlr-v37-ganin15} \cite{DBLP:journals/corr/TzengHDS15} \cite{Sener:2016:LTR:3157096.3157333} \cite{DBLP:journals/corr/TzengHZSD14} \cite{adda_cvpr2017}.
%The key idea of domain adaptation is to map both source and target domains into a common feature space.
Ganin et al. \cite{pmlr-v37-ganin15} proposed to incorporate both classification and domain adaptation to a unified network so that both tasks can be learned together.
Similarly, Tzeng et al. \cite{adda_cvpr2017} later introduced a unified framework for Unsupervised Domain Adaptation based on adversarial learning objectives (ADDA). It uses a loss function in a discriminator to be solely dependent on its target distribution. 
%{\color{red}It aims to learn  domain adaptation and classification at the same time.} {\color{blue} Hai cau nay co ve giong y qua}
%
Liu et al. \cite{NIPS2016_6544} presented Coupled Generative Adversarial Network (CoGAN) for learning a joint distribution of multi-domain images. It is then applied to domain adaptation.

\begin{table*} [!t]
	\small
	\centering
	\caption{Comparison in the properties between our proposed approaches (UNVP and E-UNVP) and other recent methods, where \xmark \HS represents \textit{not applicable} properties. Gaussian Mixture Model (GMM), Probabilistic Graphical Model (PGM), Convolutional Neural Network (CNN), Adversarial Loss ($\ell_{adv}$), Log Likelihood Loss ($\ell_{LL}$), Cycle Consistency Loss ($\ell_{cyc}$), Discrepancy Loss ($\ell_{dis}$) and Cross-Entropy Loss ($\ell_{CE}$).} 

	\begin{tabular}{ c|c|c|c|c|c|c|c}
	%c {1.65cm}}
		\hline
		& \begin{tabular}{@{}c@{}}\textbf{Domain}\\ \textbf{Modelity} \end{tabular}& \textbf{Architecture}& \begin{tabular}{@{}c@{}}\textbf{Loss}\\ \textbf{Function}\end{tabular}& %\begin{tabular}{@{}c@{}}\textbf{Model}\\ \textbf{Type} \end{tabular}
		\begin{tabular}{@{}c@{}}\textbf{End-to}\\\textbf{-End}\end{tabular} & \begin{tabular}{@{}c@{}}\textbf{Target-domain} \\ \textbf{sample-free}\end{tabular} &\begin{tabular}{@{}c@{}}\textbf{Target-domain} \\ \textbf{label-free}\end{tabular} &  \begin{tabular}{@{}c@{}}\textbf{Deployable} \\ \textbf{Domains}\end{tabular}\\
% 		\Xhline{24\arrayrulewidth}
		FT \cite{feature_transfer_learning} & Transfer Learning & CNN & $\ell_{2}$
		& \cmark & \xmark & \xmark & Two \\
		\hline
		\hline
		UBM \cite{Reynolds00speakerverification}&Adaptation & GMM & $\ell_{LL}$
		& \xmark & \xmark & \cmark & Any \\
		DANN \cite{pmlr-v37-ganin15}&Adaptation & CNN & $\ell_{adv}$
		& \cmark & \xmark & \cmark & Two \\
        CoGAN \cite{NIPS2016_6544}&Adaptation & CNN+GAN & $\ell_{adv}$
		& \cmark & \xmark & \cmark & Two \\
		I2IAdapt \cite{Murez_2018_CVPR} &Adaptation & CNN+GAN & $\ell_{adv}+\ell_{cyc}$
		& \cmark & \xmark & \cmark & Two \\
		ADDA \cite{TzengHSD17} &Adaptation & CNN+GAN & $\ell_{adv}$
		& \cmark & \xmark & \cmark & Two \\
		MCD \cite{saito2018maximum} &Adaptation & CNN+GAN & $\ell_{adv} + \ell_{dis}$
		& \cmark & \xmark & \cmark & Two \\
		\hline
		\hline
		%\Xhline{1\arrayrulewidth}
		CrossGrad \cite{shankar2018generalizing} &Generalization & Bayesian Net & $\ell_{CE}$
		& \cmark & \cmark & \cmark & Any \\
		ADA \cite{generalize-unseen-domain} &Generalization & CNN &$\ell_{CE}$
		& \cmark & \cmark & \cmark & Any \\
		\textbf{Our UNVP} &\textbf{Generalization} & \textbf{PGM+CNN} & \textbf{$\boldsymbol{\ell_{LL}} + \boldsymbol{\ell_{CE}} $}
		& \cmark & \cmark & \cmark & \textbf{Any} \\
		\textbf{Our E-UNVP} &\textbf{Generalization} & \textbf{PGM+CNN} & \textbf{$\boldsymbol{\ell_{LL}} + \boldsymbol{\ell_{CE}} $}
		& \cmark & \cmark & \cmark & \textbf{Any} \\
		%\textbf{} &\xmark &\xmark & \cmark & \cmark & \cmark & \cmark \\%& \cmark  \\
		\hline
	\end{tabular}
	
	\label{tab:TenMethodSumm}
	\vspace{-5mm}
\end{table*}

\textbf{Domain Generalization} aims to learn a classification model from a single-source domain and generalize that knowledge to achieve high performance in unseen target domains robustly. 
%A critical problem in domain generalization involves learning domain-invariant representations.  H. Li et al. proposed a novel framework for domain generalization, denoted by MMD-AAE \cite{domain_generalization_Li_2018_CVPR}.
%H. Li et al. proposed 
%{\color{blue} Rewrote this part. Check if it is correct}
To learn a domain-invariant feature representation, M. Ghifary et al. \cite{domain_generalization_Ghifary_2015_ICCV}  used multi-view autoencoders to perform cross-domain reconstructions. Later, \cite{domain_generalization_Li_2018_CVPR} introduced MMD-AAE to learn a feature representation by jointly optimizing a multi-domain autoencoder regularized via the Maximum Mean Discrepancy (MMD) distance. Recently, K. Muandet et al. \cite{domain_generalization_Muandet_ICML_2018} presented a kernel-based algorithm for minimizing the differences in the marginal distributions of multiple domains, whereas Y. Li \cite{domain_generalization_Li_2018_ECCV} proposed an end-to-end conditional invariant deep domain generalization approach by leveraging deep neural networks for domain-invariant representation learning. To address the problem of unseen domains, R. Volpi et al. presented Adversarial Data Augmentation (ADA) \cite{generalize-unseen-domain} to generalize to unseen domains.

\begin{figure}[t]
	\centering \includegraphics[width=1.0\columnwidth]{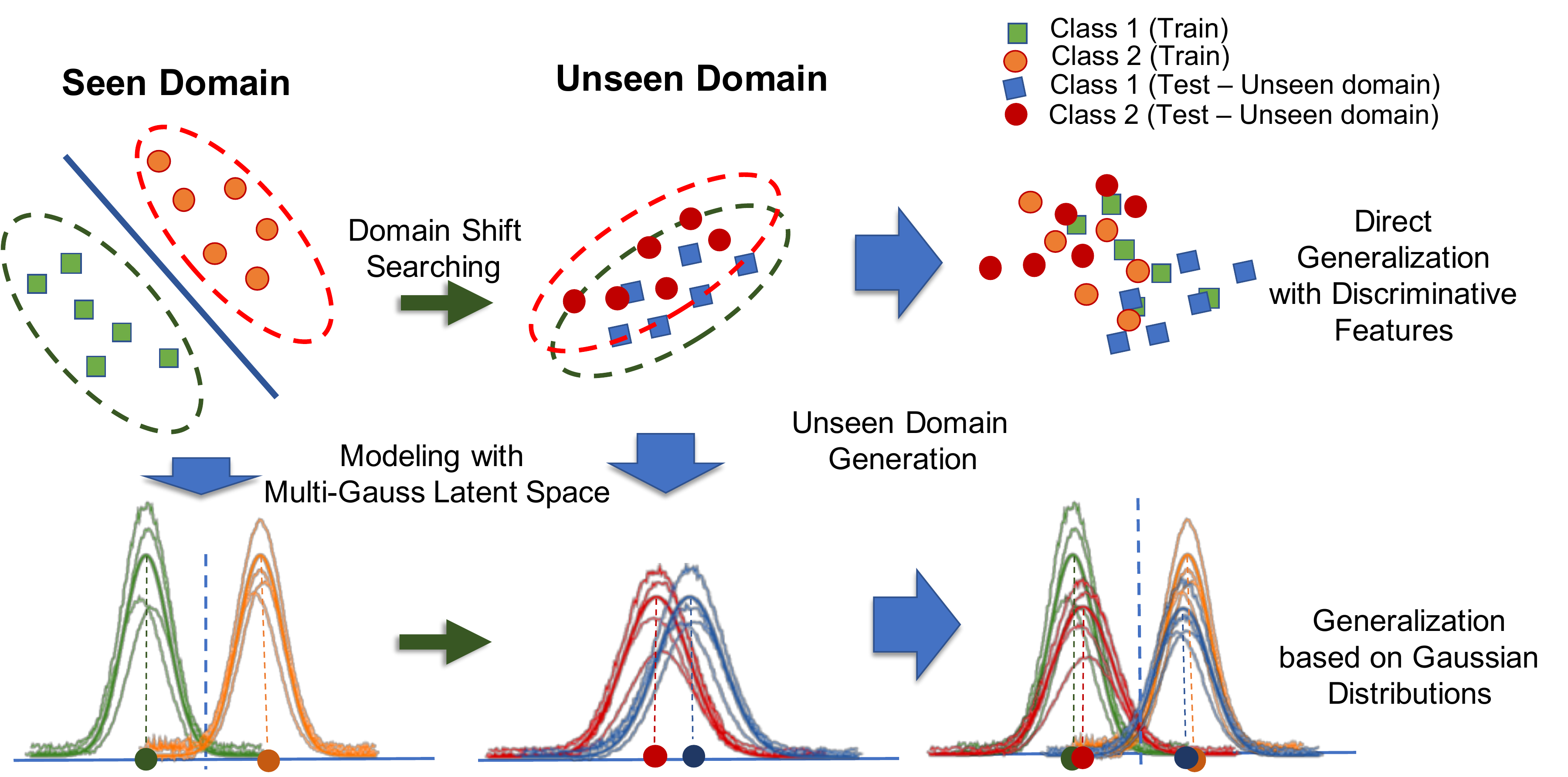}
	\caption{Illustration of the proposed UNVP method. The traditional classifier fails to model new samples in unseen domains (top). Meanwhile, UNVP consistently maintains the feature distribution in each class while searching for a new shifting domain (bottom).}	
	\label{fig:domain_shift}
\end{figure}

\section{The Proposed Method} \label{sec:proposed_method}

Far apart from previous augmentation methods that tried to generate new  %{\color{red}{
samples in image space
%}} {\color{blue}{Y nay chua ro lam. Do la new unseen samples?}} 
using prior knowledge with the hope that these samples can cover unseen domains, our approach, on the other hand, focuses on modeling the environment density as multiple Gaussian distributions in a deep feature space and uses this knowledge for the generalization process. In this way, the new samples are automatically synthesized with more semantic meaning while consistently maintaining the feature structures (see Fig. \ref{fig:domain_shift}). Thus, without the need to see the samples in target domains, the method is still able to handle the domain shifting effectively and robustly achieves high performance in these unseen domains.

%In our approach, new samples are synthesized {\color{red}{to expand these distributions}} 
%{\color{blue}{Expand the distributions co ve khong dung lam. CO the tim tu phu hop }} 
%so that they can effectively handle the domain shifting and  {\color{red}{cover the distributions of samples}} {\color{blue}{tai sao la cover & tai sao the distributions (day la distributions of unseen samples?}} in unseen domains as much as possible.
%our approach \textit{focuses on augmentation in semantic deep feature space} by \textit{estimating of environment density}. By this way, more semantic samples can be synthesized and, therefore, generalize the learning process.

In particular, the proposed UNVP and E-UNVP approaches present a new \textit{tractable CNN deep network} to extract the deep features of samples in the source environment and formulate their probability densities to \textit{multiple Gaussian} distributions (Fig. \ref{fig:domain_shift}).  From these learned distributions, a density-based augmentation approach is employed to expand data distributions of the source environment for generalizing to different unseen domains. 
This architecture design allows unifying deep feature modeling and distribution modeling within an end-to-end framework.
% as in Fig. \ref{fig:Method}. 

The proposed framework consists of two main streams: (1) Discriminative feature modeling with a deep network classifier; and (2) Deep Generative Flows to model the domain variations in the form of distributions. %{\color{red}distributions}{\color{blue} Is that Gaussian Distributions}. 
They are together going through an end-to-end learning process that alternatively minimizes the within-class distributions and synthesizing new useful samples to generalize to new unseen domains.
%(1) \textit{Domain variation modeling} via deep mapping functions; (2) \textit{Unseen domain generalization}; and (3) \textit{End-to-end joint training deep network}. 
Notice that our proposed framework does not require the presence of samples in the target domains during the training process. %{\color{blue} check if it is correct}
%In the next sub-sections, further details of the proposed Deep Generative Flows 

%In the next subsections, we will introduce the deep generative flow to model samples in the source domain as Gaussian distributions according to their class labels (Sec. \ref{sec:DensityModeling}), and how to expand these distributions to unseen domain (Sec. \ref{sec:DomainGeneralization}). Finally, two variants of UDGF are introduced in Sec. \ref{sec:UDGF}.

\subsection{Domain Variation Modeling as Distributions}
\label{sec:DensityModeling}
%Modeling environment variation directly in high-dimensional image domain is extremely complicated and easy to diverse due to the effects of noisy samples.
This section aims at learning a Deep Generative Flow model, i.e. function $\mathcal{F}$, that maps an image $\mathbf{x}$ in image space $\mathcal{I}$ to its latent representation $\mathbf{z}$ in latent domain $\mathcal{Z}$ such that the density function $p_X(\mathbf{x})$ can be estimated via 
the probability density function $p_Z(\mathbf{z})$.
Then via $\mathcal{F}$, rather than representing the environment variation, i.e. $p_X(\mathbf{x})$, directly in the image space, %{\color{red}it} {\color{blue} Is it the given image x. If so, please replace it by the given image x}
it can be easily modeled via variables in latent space, i.e. $p_Z(\mathbf{z})$, with more semantic manner. When $p_Z(\mathbf{z})$ follows prior distributions, all samples in the given domain can be effectively modeled in the forms of latent distributions.

\noindent
\textit{\textbf{Structure and Variable Relationship.}}
Let $\mathbf{x} \in \mathcal{I}$ be a data sample in image domain $\mathcal{I}$, $y$ be its corresponding class label, and $\mathbf{z} = \mathcal{F}(\mathbf{x}, y,\theta)$ where $\theta$ denotes the parameters of $\mathcal{F}$, the probability density function of $\mathbf{x}$ can be formulated via the change of variable formula as follows:
\begin{equation} \label{eqn:DensityFunc}
\small
    p_X(\mathbf{x},y;\theta) = p_Z(\mathbf{z},y;\theta)\left|\frac{\partial \mathcal{F} (\mathbf{z},y;\theta)}{\partial\mathbf{x}} \right|
\end{equation}
where $p_X(\mathbf{x},y)$ and $p_Z(\mathbf{z},y;\theta)$ define the distributions of samples of class $y$ in image and latent domains, respectively. $\frac{\partial \mathcal{F} (\mathbf{z}, y;\theta)}{\partial\mathbf{x}}$ denotes the Jacobian matrix with respect to $\mathbf{x}$. Then the log-likelihood is computed by.

\begin{equation} \label{eqn:LogLikelihoodFunc}
\small
    \log p_X(\mathbf{x},y;\theta) = \log p_Z(\mathbf{z},y;\theta) + \log\left|\frac{\partial \mathcal{F} (\mathbf{z},y;\theta)}{\partial\mathbf{x}} \right|
\end{equation}
Eqns. \eqref{eqn:DensityFunc} and \eqref{eqn:LogLikelihoodFunc} provide two facts: (1) learning the density function of samples in class $y$ is equivalent to estimate the density of its latent representation $\mathbf{z}$ and determinant of the associated Jacobian matrix $\frac{\partial \mathcal{F}}{\partial \mathbf{x}}$; and (2) if the latent distribution $p_Z$ is defined as a Gaussian distribution, the learned function $\mathcal{F}$ explicitly becomes the mapping function from a real data distribution to a Gaussian distribution in latent space. 
Then, we can model the environment variation via deviations from the Gaussian distributions of all classes in a latent domain.
When $\mathcal{F}$ is well-defined with tractable computation of its Jacobian determinant, the two-way connection, i.e., inference and generation, is existed for $\mathbf{x}$ and $\mathbf{z}$. 
%Motivated from these properties, in order to learn the mapping function $\mathcal{F}$ and compute density function $p_X$, the 

\noindent
\textbf{\textit{The prior class distributions.}} Motivated from these properties, given $C$ classes, we choose $C$ Gaussian distributions with different means $\{\boldsymbol{\mu}_1, \boldsymbol{\mu}_2, .., \boldsymbol{\mu}_C\}$ and covariances $\{\Sigma_1,\Sigma_2,...,\Sigma_C\}$ as prior distributions for these classes, i.e. $\mathbf{z}_c \sim \mathcal{N}(\boldsymbol{\mu}_c, \Sigma_c)$. 
It is worth noting that even when we choose Gaussian Distributions, our framework is not limited to other distribution types.

% \begin{figure}[!t]
% 	\centering \includegraphics[width=0.8\columnwidth]{Figures/framework.png}
% 	\caption{The Proposed Method.}
% 	\label{fig:Method}
% \end{figure}

\noindent
\textit{\textbf{Mapping function structure.}}
To enforce the information flow from an image domain to a latent space with different abstraction levels, we formulate the mapping function $\mathcal{F}$ as a composition of several sub-functions $f_i$ as follows.
\begin{equation} \label{eqn:MappingFunction}
\small
    \mathcal{F} = f_1 \circ f_2 \circ ... \circ f_N
\end{equation}
where $N$ is the number of sub-functions. The Jacobian $\frac{\partial \mathcal{F}}{\partial \mathbf{x}}$ can be derived by $\frac{\partial \mathcal{F}}{\partial \mathbf{x}} = \frac{\partial f_1}{\partial \mathbf{x}} \cdot \frac{\partial f_2}{ \partial f_1} \cdots \frac{\partial f_N}{ \partial f_{N-1}}$. With this structure, the properties of each $f_i$ will define the properties for the whole mapping function $\mathcal{F}$. For example, if the Jacobian of $\frac{\partial f_1}{\partial \mathbf{x}}$ is tractable, then $\mathcal{F}$ is also tractable. Furthermore, if $f_i$ is a non-linear function built from a composition of CNN layers then $\mathcal{F}$ becomes a deep convolution neural network. There are several ways to construct the sub-functions, i.e. different CNN structures for non-linearity property. 
%In our approach, the sub-function in \cite{Duong_2017_ICCV} is adopted thanks to its tractable and invertible.
%
\begin{equation} \label{eqn:MappingUnit}
\small
    f(\mathbf{x}) = \mathbf{b} \odot \mathbf{x} + (1-\mathbf{b) \odot \left[\mathbf{x} \odot \exp{(\mathcal{S}(\mathbf{b} \odot \mathbf{x})} + \mathcal{T}(\mathbf{b} \odot \mathbf{x})\right]}
\end{equation}
where $\mathbf{b}=[1,...,1,0,...,0] $ is a binary mask,
%{\color{blue} what is binary mask and how to obtain it}
and $\odot$ is the Hadamard product. $\mathcal{S}$ and $\mathcal{T}$ define the scale and translation functions during mapping process.

\noindent
\textit{\textbf{Learning the mapping function and Environment Modeling.}}
To learn the parameter $\theta$ for mapping function $\mathcal{F}$, the log-likelihood in Eqn. \eqref{eqn:LogLikelihoodFunc} is maximized as follows.
\begin{equation}
\small
    \theta^* = \arg \max_{\theta} \sum_{c}\sum_{i}\log p_X(\mathbf{x}^i,c;\theta)
\end{equation}

\begin{figure}[t]
 	\centering \includegraphics[width=0.8\columnwidth]{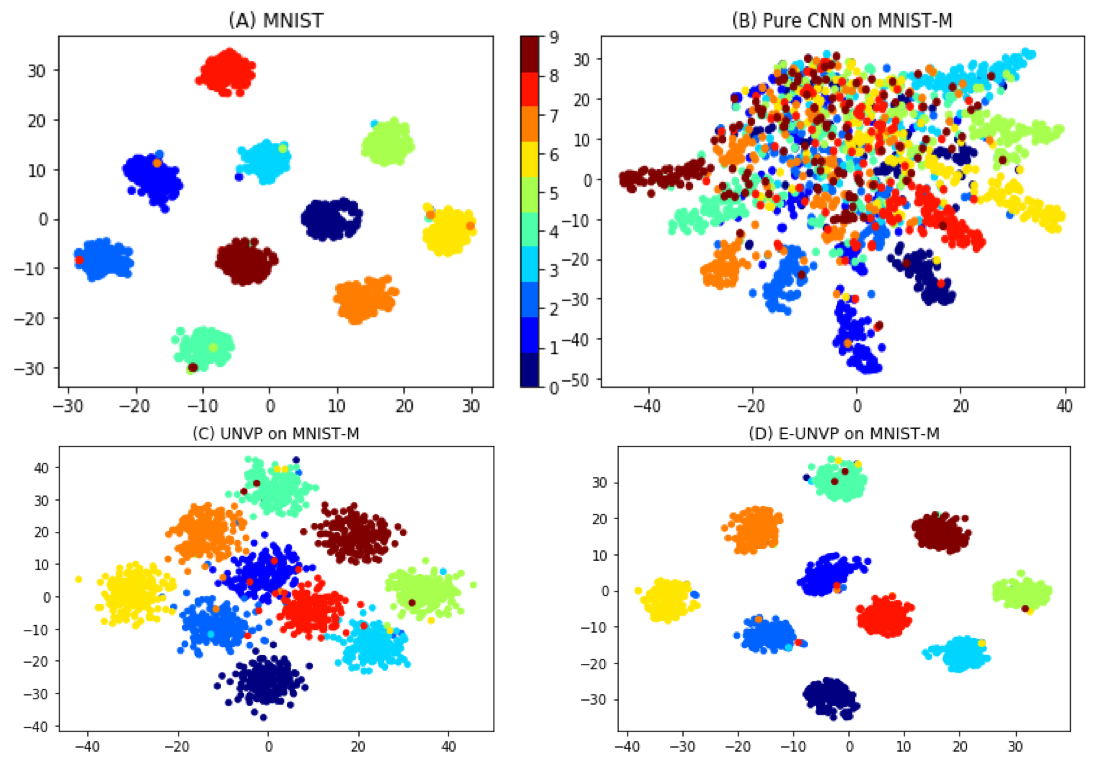}
%  	\caption{The distributions: (A) The left figure is the distribution of MNIST dataset. The right figure is the distributions of MNIST and MNIST-M datasets. The right image illustrates that when modeling MNIST-M using the model trained on MNIST, the distribution of MNIST-M is shifted to the right. (B) The distributions of class 0 and 8 of MNIST and MNIST-M after using unseen domain generalization to train environment variation modeling.}
    \caption{The distributions: (A) MNIST. (B) MNIST-M using a Pure-CNN trained on MNIST, (C) MNIST-M using our UNVP trained on MNIST. (D) MNIST-M using our E-UNVP trained on MNIST.}
 	\label{fig:Distribution}
 \end{figure}

Notice that after learning the mapping function, \textbf{\textit{all images of all classes are mapped into the corresponding distributions of their classes.}} Then the environment density can be considered as the composition of these distributions. Figure \ref{fig:Distribution}(A) illustrated an example of the learned environment distributions of MNIST with 10 digit classes. When only samples in MNIST are used for training, the density distributions of MNIST-M, i.e., \textit{unseen during training}, using Pure-CNN, in our UNVP and E-UNVP are shown in Fig. \ref{fig:Distribution} (B, C, D), respectively.
%two different environments (i.e. MNIST and MNIST-M) are also presented in Figure \ref{fig:Distribution}(A) (right). 
In the next section, a generalization approach is proposed so that using only samples in a source environment, the learned model can expand the density distributions of the source environment so that they can cover as much as possible the distributions of unseen environments.

\subsection{Unseen Domain Generalization} \label{sec:DomainGeneralization}

After modeling the source environment variation as the compositions of its class distributions, this section introduces the generalization process of these distributions with respect to a classification model $\mathcal{M}$ such that the expansion of these distributions can help $\mathcal{M}$ generalize to unseen environments with high accuracy. Notice that $\mathcal{M}$ can be any type of Deep CNN such as LeNet \cite{lenet_ref}, AlexNet \cite{deep_alexnet}, VGG \cite{deep_vgg}, ResNet \cite{deep_resnet}, DenseNet \cite{deep_densenet}. 

Let $\ell(\mathbf{X,Y};\mathcal{M},\mathcal{F},\theta, \theta_1)$ be the training loss function of $\mathcal{M}$, and $\theta_1$ be the parameters of $\mathcal{M}$. The generalization process of $\mathcal{M}$ can be formulated as updating the parameters $\theta_1$ such that it can robustly classify the samples having latent distributions that are distance $\rho$ away from the samples in the source environment.
%
%even the class distributions of the unseen environment are distance $\rho$ away from the source environment, $\mathcal{M}$ is still robust with high accuracy.
Then, the objective function of $\mathcal{M}$ is formulated as.
\begin{equation} \label{Eqn:WorseCaseFormulation}
\small
    \arg \min_{\theta_1} \sup_{P:d(P_X,P_X^{src})\leq \rho} \mathbb{E}\left[ \ell(\mathbf{X,Y};\mathcal{M},\mathcal{F},\theta, \theta_1)\right] 
\end{equation}
where $\{\mathbf{X,Y}\}$ denotes the images and their labels; $d(\cdot, \cdot)$ is the distance between probability distributions; $P^{src}_X(\mathbf{X,Y})$ and $P_X(\mathbf{X,Y})$ are the density distributions of the source and current expanded environments, respectively.

Since both $P_X^{src}$ and $P_X$ are density distributions, the Wasserstein distance with respect to $P_X^{src}$ and $P_X$ can be adopted.
% as follows.
%
% \begin{equation}
%     d(P_X,P_X^{src}) = \sum_{c} \sum_{\mathbf{x}_c,\mathbf{x}_c^{src}} \inf \mathbb{E} \left[ cost\left( \mathbf{x}_c,\mathbf{x}_c^{src}\right) \right]
% \end{equation}
% \noindent
% where $cost(\cdot, \cdot)$ denotes the transformation cost. 
Notice that from previous section, we have leaned a mapping function $\mathcal{F}$ that maps the density functions from image space, i.e. $P_X$, to prior distributions in latent space, i.e. $P_Z$. Moreover, since $\mathcal{F}$ is invertible with the specific formula of its sub-functions, computing $d(P_X,P_X^{src})$ is equivalent to $d(P_Z,P_Z^{src})$. From this, we can efficiently estimate $cost$ as the transformation cost between Gaussian distributions. 
Then $d(P_X,P_X^{src})$ is reformulated by.
\begin{equation} \label{eqn:DistanceW}
\small 
\begin{split}
    d(P_X,P_X^{src}) &= d(P_Z,P_Z^{src})\\
    & = \sum_{c}\sum_{\mathbf{x}_c,\mathbf{x}_c^{src}} \inf \mathbb{E} \left[ cost\left( \mathcal{F}(\mathbf{x}_c),\mathcal{F}(\mathbf{x}_c^{src})\right) \right]\\
    & = \sum_{c}\sum_{\mathbf{z}_c,\mathbf{z}_c^{src}} \inf \mathbb{E} \left[ cost\left( \mathbf{z}_c,\mathbf{z}_c^{src}\right) \right]
\end{split}    
\end{equation}
where $cost(\cdot, \cdot)$ denotes the transformation cost between Gaussian distributions:
\begin{equation} \label{eqn:DistanceWCost}
\small 
\begin{split}
%    cost^2\left( \mathbf{z}_c,\mathbf{z}_c^{src}\right)&= || \mu{'}_c - \mu_c ||^2_2 \\
%    & + \text{Tr}(\Sigma'_c + \Sigma_c - 2(\Sigma'^{1/2}_c\Sigma_c\Sigma'^{1/2}_c)^{1/2})
cost^2(\mathbf{z}_c,\mathbf{z}_c^{src})
    =&\sum_c || \mu^{src}_c - \mu_c ||^2_2 \\
    + \text{Tr}(\Sigma^{src}_c &+ \Sigma_c - 2((\Sigma^{src}_c)^{1/2}\Sigma_c(\Sigma_c^{src})^{1/2})^{1/2})
\end{split}    
\end{equation}
\noindent
$\{\mu_c, \Sigma_c\}$ and $\{\mu'_c, \Sigma'_c\}$ are the means and covariances of the distributions of class $c$ in the source and the expanded environment, respectively.
Plugging this distance and applying the Lagrangian relaxation to Eqn. \eqref{Eqn:WorseCaseFormulation}, we have
\begin{equation} \label{eqn:argsol} \nonumber
\small 
\begin{split}
    &\arg \min_{\theta_1} \sup_P \mathbb{E} \left[\ell(\mathbf{X,Y};\mathcal{M},\mathcal{F}, \theta, \theta_1)\right] - \alpha \cdot d(P_X,P_X^{src})\\
    =&\arg \min_{\theta_1} \sum_c \sup_{\mathbf{x}} \{ \ell(\mathbf{x},c;\mathcal{M},\mathcal{F}, \theta, \theta_1) - 
    \alpha \cdot cost(\mathbf{\mathcal{F}(x)},\mathcal{F}(\mathbf{x}_c^{src})) \}
\end{split}
\end{equation}
To solve this objective function, the optimization process can be divided into two alternative steps: (1) generate the sample $\mathbf{x}$ for each class such that 
\begin{equation} \label{eqn:GenerateNewSample}
\small
    \mathbf{x}=\arg \max_{\mathbf{x}} \{ \ell(\mathbf{x},c;\mathcal{M},\mathcal{F}, \theta, \theta_1) - \alpha \cdot cost(\mathbf{\mathcal{F}(x)},\mathcal{F}(\mathbf{x}_c^{src})) \}
\end{equation}
and consider $\mathbf{x}$ as a new ``hard'' example for class $c$; and (2) add $\mathbf{x}$ to the training data and optimize the model $\mathcal{M}$. In other words, this two-step optimization process aims at finding new samples belonging to distributions that are $\rho$ distance far away from the distributions of the source environment, and making $\mathcal{M}$ became more robust when classifying these examples. 
In this way, after a certain of iteration, the distributions learned from $\mathcal{M}$ can be generalized so that they can cover as much as possible the distributions of new unseen environments.
\begin{figure}[!t]
	\centering \includegraphics[width=1.0\columnwidth]{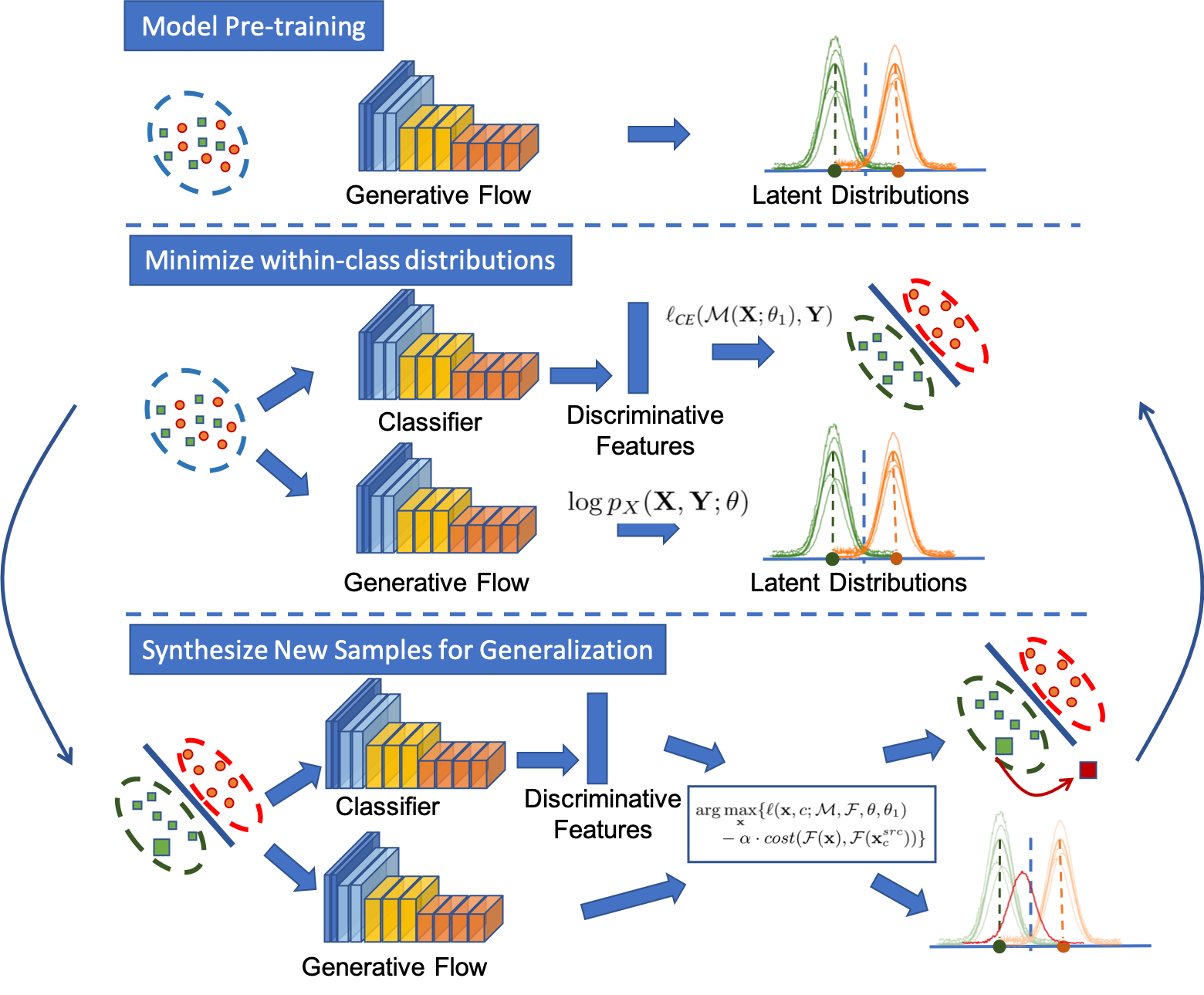}
	\caption{\textbf{The Training Process of Our proposed UNVP.} consists of one pre-training step and a two-stage optimization by alternatively minimizing the within-class distributions and synthesizing new samples for generalization.}	
	\label{fig:TrainingProcess}
\end{figure}
\subsection{Universal Non-volume Preserving (UNVP) Models} \label{sec:UDGF}
The proposed UNVP consists of two main branches: (1) \textit{Discriminative} Feature Modeling and (2) \textit{Generative} Distribution Modeling. While the discriminative part focuses on constructing a classifier that minimizes within-class distributions, the generative one aims at embedding samples of all classes into their corresponding latent distributions and then expanding these distributions for generalization.
Fig. \ref{fig:TrainingProcess}  illustrates the whole end-to-end joint training process for UNVP where the generative part, i.e., \textit{Deep Generative Flow} $\mathcal{F}$, is firstly employed to learn the mapping from image space to Gaussian distributions in latent space. Then a two-stage training process is adopted
%including (1) optimize the classifier $\mathcal{M}$ and deep mapping function $\mathcal{F}$ 
to learn the Deep Classifier $\mathcal{M}$ and adjust the Deep Generative Flow $\mathcal{F}$ for generalization. 

In the first stage of this process, given a training dataset, both parameters $\{\theta, \theta_1\}$ of the mapping function $\mathcal{F}$ and the classifier $\mathcal{M}$ are updated according to the loss function as.
\begin{equation*}
\small
\begin{split}
    \ell(\mathbf{X,Y};\mathcal{M},\mathcal{F},\theta, \theta_1) = &\ell_{\text{\textit{CE}}}( \mathcal{M}(\mathbf{X};\theta_1),\mathbf{Y} - \log p_X(\mathbf{X},\mathbf{Y}; \theta)
\end{split}
\end{equation*}
where the first term is the cross-entropy loss for $\mathcal{M}$ and the second term is the log-likelihood of $\mathcal{F}$.

In the second stage, we adapt the generalization process as presented in Sec. \ref{sec:DomainGeneralization} and Eqn. \eqref{eqn:GenerateNewSample} to synthesize new ``hard'' samples. Notice that, to further constraint the perturbation in latent space, we incorporate another regularization term to Eqn. \eqref{eqn:DistanceW} as.
\begin{equation} \nonumber
\small 
\begin{split}
    cost^2(\mathbf{z}_c,\mathbf{z}_c^{src})
    =&\sum_c || \mu^{src}_c - \mu_c ||^2_2 \\
    & + \text{Tr}(\Sigma^{src}_c + \Sigma_c - 2((\Sigma^{src}_c)^{1/2}\Sigma_c(\Sigma_c^{src})^{1/2})^{1/2}) \\
    & + || \mathcal{M}(\mathbf{X}_c) - \mathcal{M}(\mathbf{X}^{src}_c)||_2^2
\end{split}    
\end{equation}
New generated samples are then added to the training set and used for updating both branches of UNVP. 

% The proposed Universal Deep Generative Flow framework is illustrated in Fig. \ref{fig:Method}.
% There are three main steps. (1) Environment Modeling (2) ... (3) ...
% The classifier learning and Domain Generalization are alternatively taken place.
Notice that in the structure of $\mathcal{F}$, the choice of Gaussian distributions for all classes play an important role and directly affects the performance of the generative model. By varying the choices for these distributions, different variants of UNVP can be introduced.
\vspace{-6pt}
\paragraph{Universal Non-volume Preserving Models (UNVP):} The means and covariances of Gaussian distributions are pre-defined for all $C$ classes where $\boldsymbol{\mu}_c = \text{\textbf{1}}c; \boldsymbol{\Sigma} = \mathbf{I}$; $\mathbf{z}_c \sim \mathcal{N}(\boldsymbol{\mu}_c, \mathbf{I})$ where \text{\textbf{1}} is the all-one vector.
\vspace{-6pt}
\paragraph{Extended Universal Non-volume Preserving Models (E-UNVP):} 
    %use learned prior distributions. we further introduce an extended version of UNVP, named E-UNVP, in the capability of learnable prior distribution for every class. In other words, 
    Rather than fixing the means and covariances of the Gaussian distributions of $C$ classes, we consider them as variables and flexibly learned during the environment modeling as well as adjusted during domain generalization. Particularly, given the class label $c$, 
%Particularly, during the Environment Modeling stages, 
$\mathcal{F}$ maps each sample $\mathbf{x_c}$ 
%in the source environment with class label $c$ 
to a Gaussian distribution with the mean and covariance as.
\begin{equation} \label{eqn:mean_sig}
\small
\begin{split}
    \boldsymbol{\mu}_c & = \gamma \mathcal{G}_m(c) + \lambda \mathcal{H}_m(\mathbf{n})\\
    \mathbf{\Sigma}_c & = \mathcal{G}_{std}(c)
\end{split}
\end{equation}
where $\mathcal{G}_m(c)$ and $\mathcal{G}_{std}(c)$ denote the learnable function that map label $c$ to the mean and covariance values of its Gaussian distribution. 
$\mathbf{n}$ is a noise signal that is generated following the normal distribution. $\mathcal{H}_m(\mathbf{n})$ defines the allowable shifting range of the Gaussian given the noise signal $\mathbf{n}$. $\gamma$ and $\lambda$ are user-defined parameters that control the separation of the Gaussian Distributions between different classes and the contribution of $\mathcal{H}_m(\mathbf{n})$ to $\boldsymbol{\mu}_c$.
We choose the Fully Connected structure for $\mathcal{G}_m(c)$ and $\mathcal{G}_{std}(c)$ that take the input $c$ in the form of one-hot vector while Convolutional Layer is adopted for $\mathcal{H}_m(\mathbf{n})$.

\section{Discussion}
As shown in Fig. \ref{fig:domain_shift}, by exploiting the Generative Flows that model samples of each class as a Gaussian in semantic feature space, the proposed UNVP can robustly maintain the feature structure of each class while expanding and shifting the domain distributions. In this way, we can generate more useful ``hard'' samples for the generalization process.

By introducing the noise signal $\mathbf{n}$, we allow the Gaussian distribution of each class shifting around within a limited range, i.e., $\mathcal{H}_m(\mathbf{n})$. This further enhances the robustness of E-UNVP against noise during the environment modeling.

To further enhance the capability of modeling the input signal with high-resolution, we incorporate the activation normalization and invertible $1 \times 1$ convolution operators \cite{glow_ref} to the structure of each sub-function $f_i$ in Eqn. \eqref{eqn:MappingFunction}. Particularly, the input to each $f_i$ is passed through an actnorm layer followed by an invertible $1 \times 1$ convolution before being transformed by $\mathcal{S}$ and $\mathcal{T}$ as in Eqn. \eqref{eqn:MappingUnit}.
The two transformations $\mathcal{S}$ and $\mathcal{T}$ are defined by two Residual Networks with rectifier non-linearity and skip connections. Each of them contains three residual blocks. For input image with the resolution higher than $128 \times 128$, six residual blocks are set for $\mathcal{S}$ and $\mathcal{T}$.
%we incorporate the actnorm and invertible $1 \times 1$ convolution operators \cite{glow_ref} before going through two residual blocks in the structure of each unit $f$.

\section{Experiments} \label{sec:experiemts}

\begin{figure}[!b]
	\centering \includegraphics[width=0.9\columnwidth]{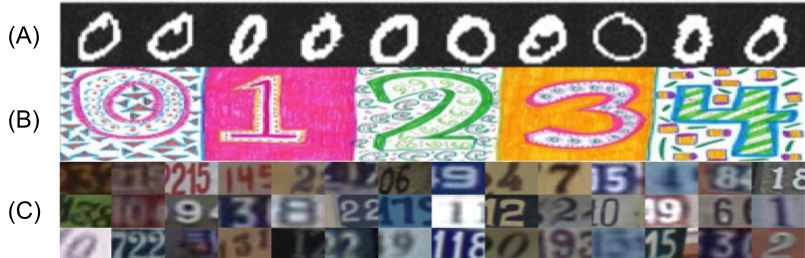}
	\caption{Examples in (A) MNIST, (B) MNIST-M, and (C) SVHN databases}
	\label{fig:char_dbs}
\end{figure}

\begin{table*}[!t]
    \small
    \centering
    \caption{Ablative experiment results (\%) on the effectiveness of the parameters $\lambda$, $\alpha$
    and $\beta$ that control the distribution separation and shitting range. MNIST is used as the only training set, MNIST-M is used as the unseen testing set.}
    %% mean = class * 10 + alpha * random_z
    \label{tab:ablation}
    \begin{tabular}{|c|c|c|c|c||c|c|c||c|c|c|c|c|}
    \hline
     \multirow{2}{*}{\textbf{Dataset}}& \multirow{2}{*}{\textbf{Methods}} & \multicolumn{3}{c||}{\boldsymbol{$\lambda$}} & \multicolumn{3}{c||}{\boldsymbol{$\alpha$}} & \multicolumn{5}{c|}{\boldsymbol{$\beta (\%)$}} \\
     \cline{3-13}
     \multirow{2}{*}{}& \multirow{2}{*}{} & $\textbf{0.01}$ & $\textbf{0.1}$ & $\textbf{1.0}$ & $\textbf{0.01}$ & $\textbf{0.1}$ & $\textbf{1.0}$ & $\textbf{0\%}$ & $\textbf{1\%}$ & $\textbf{10\%}$ & $\textbf{20\%}$ & $\textbf{30\%}$ \\
    \hline
     \multirow{3}{*}{MNIST}     & {Pure-CNN}        & \multicolumn{11}{c|}{99.28} \\ 
        %& 99.28 & 99.28 & 99.28 & 99.28 & 99.28 & 99.28 & 99.28 & 99.28 & 99.28 & 99.28 & 99.28 \\
     \cline{3-13}
     \multirow{3}{*}{}          & \textbf{UNVP}     & $-$    & $-$    & $-$     & \textbf{99.33} & 99.18 & 99.30 & 99.28 & 99.28 & \textbf{99.35} & 99.30 & 99.36 \\
     \cline{3-13}
     \multirow{3}{*}{}          & \textbf{E-UNVP}   & \textbf{99.22} & \textbf{99.42} & \textbf{99.40}  & 99.13 & \textbf{99.31} & \textbf{99.42} & \textbf{99.28} & \textbf{99.36} & 99.34 & \textbf{99.42} & \textbf{99.43} \\
     \hline
     \multirow{3}{*}{MNIST-M}   & {Pure-CNN}        & \multicolumn{11}{c|}{55.90} \\ 
     %xx.xx & xx.xx & xx.xx & xx.xx & xx.xx & xx.xx & 58.74 & xx.xx & xx.xx & 58.74 & xx.xx \\
     \cline{3-13}
     \multirow{3}{*}{}          & \textbf{UNVP}     & $-$    & $-$    & $-$    & \textbf{58.18} & 60.76 & 59.44 & 55.90 & \textbf{59.99} & 57.24 & 59.44 & 55.11 \\
     \cline{3-13}
     \multirow{3}{*}{}          & \textbf{E-UNVP}   & \textbf{59.83} & \textbf{60.49} & \textbf{59.47} & 56.92 & \textbf{61.70} & \textbf{60.49} & \textbf{55.90} & 57.10 & \textbf{60.49} & \textbf{61.70} & \textbf{60.49} \\
     \hline
\end{tabular}
\end{table*}

 This section first shows the effectiveness of our proposed methods with comprehensive ablative experiments. 
In these experiments, we use MNIST as \textit{the only} the training set and MNIST-M as the unseen testing set. 
The proposed approaches are also benchmarked on various deep network structures, i.e. LeNet \cite{lenet_ref}, AlexNet \cite{deep_alexnet}, VGG \cite{deep_vgg}, ResNet \cite{deep_resnet} and DenseNet \cite{deep_densenet}.
Using the final optimal model, we show in the next subsection that our approaches consistently achieve the state-of-the-art results in digit recognition on three-digit datasets, i.e., MNIST, SVHN \cite{svhn_dataset}, and MNIST-M.
Then, we show the results of our proposed approaches in face recognition in three databases, i.e. Extended Yale-B \cite{yale_b_dataset}, CMU-PIE \cite{pie_dataset} and CMU-MPIE \cite{multi_pie_dataset}. 
We use facial images with normal illumination as the training domain and the ones in dark illumination conditions as the testing set on the new unseen domains.
Finally, we show the advantages of UNVP and E-UNVP in the cross-domain pedestrian recognition on the Thermal Database.

\subsection{Ablation Study}
\label{sec:ablation}

%Params:
%Scale (D)
%Single Gauss v.s Multi Gauss
%NVP: single Gass: mean + STD (fixed: 0, 1)
%NVP: multi Gass 
%mean + STD (fixed: 0, 1): Scale =  class-label * scale
%Glow: single Gass: mean + STD (learned: conv + FC): image 32x32, conv: 3x3x96x96 (48: feature size (mean (48) + std (48))

This experiment aims to measure the effectiveness of the domain generalization and perturbation processes 
% (Fig. \ref{fig:Method}).
This experiment uses MNIST as the only training set and MNIST-M as the testing one. %Testing samples in MNIST-M are generated by blending digits from the original set over patches randomly extracted from color photos from BSDS500 \cite{countour_detection}.
To simplify the experiment, LeNet \cite{lenet_ref} is used as the classifier, i.e., Pure-CNN.
% This deep network can be technically replaced by any other deep network models. We use a CNN with the designed architecture \textit{conv-pool-conv-pool-fc-fc-softmax}.
About the network hyper-parameters, we choose the learning rate and the batch size to 0.0001 and 128, respectively. 

% We do $K$ times of the maximize phase, each time will randomly select $\beta$ percent of training images to generate new hard examples . Adam Optimizer is used to optimize the deep network.

\noindent \textbf{Hyper-parameter Settings.} In the GLOW learning process, the multiple Gaussian distributions are handled via the set of scale parameters, i.e., $\gamma$ and $\lambda$, to control the distribution separation and shitting range as in Eqn. (\ref{eqn:mean_sig}).
The contributions of the generalization process are also evaluated with various percentages of ``hard'' generated samples ($\beta$), i.e., from $0\%$ to $30\%$. When $\beta = 0$, there are no new samples. 
%When $\beta = 50\%$, perturbed samples are generated as a half of the original training data.

%We adopt Real-NVP \cite{real_nvp} and GLOW \cite{glow_ref} as UNVP and E-UNVP, respectively. To generate hard examples, we use a framework of Adversarial Data Augmentation \cite{generalize-unseen-domain}. 
There are two phases alternatively updated in the training process: (1) Minimization phase to optimize the networks
and (2) Maximization (perturb) phase to generate new hard examples. We do $K$ times of the maximization phase, for each time, we randomly select $\beta$ percent of the number of training images to generate new hard samples via deep generative models. Specifically, our maximization phase generalizes new images based on both semantic features from the CNN classifier and the semantic space via the estimation of environment density.
The experimental results in Table \ref{tab:ablation} show that the proposed approaches consistently help to improve the classifiers.

\noindent \textbf{Sample Distributions in Unseen Domains.} The sample class distributions with the optimal parameter set are used to visually observed and demonstrated in Fig. \ref{fig:Distribution}. While Pure-CNN obviously fails to model unseen domain MNIST-M dataset, our UNVP successfully does domain shift and cover unseen domain dataset. These sample distributions are completely class separated when using our E-UNVP. 
% \footnote{See supplementary for the distributions in other databases}.

\begin{table}[!b]
    \small
    \centering
    \caption{Experimental results  $(\%)$ when using UNVP and E-UNVP in various common CNNs.}
    \label{tab:networks}
    
    \begin{tabular}{|c|c|c|c|}
        \hline
        \textbf{Networks} & \textbf{Methods} & \textbf{MNIST} & \textbf{MNIST-M} \\
        \hline
        \multirow{3}{*}{LeNet}      & Pure-CNN          & 99.06  & 55.90 \\
        \cline{2-4}
        \multirow{3}{*}{}           &  \textbf{UNVP}    & 99.30 & 59.44 \\
        \cline{2-4}
        \multirow{2}{*}{}           &  \textbf{E-UNVP}  & \textbf{99.42} & \textbf{61.70} \\
        \hline
        
        \multirow{3}{*}{AlexNet}    & Pure CNN          & \textbf{99.17}  & 40.12 \\
        \cline{2-4}
        \multirow{3}{*}{}           &  \textbf{UNVP}    & 98.81  & 39.94 \\
        \cline{2-4}
        \multirow{3}{*}{}           &  \textbf{E-UNVP}  & 98.89  & \textbf{40.60} \\
        \hline
        
        \multirow{3}{*}{VGG}        & Pure CNN & 99.43 & 50.67 \\
        \cline{2-4}
        \multirow{3}{*}{}           &  \textbf{UNVP}    & \textbf{99.42}  & \textbf{54.41} \\
        \cline{2-4}
        \multirow{3}{*}{}           &  \textbf{E-UNVP}  &  99.40 & 51.37 \\
        \hline
        
        \multirow{3}{*}{ResNet}     & Pure CNN          & 98.01  & 35.35 \\
        \cline{2-4}
        \multirow{3}{*}{}           &  \textbf{UNVP}    &  98.82 & 37.15 \\
        \cline{2-4}
        \multirow{3}{*}{}           &  \textbf{E-UNVP}  &  \textbf{98.97} & \textbf{40.60} \\
        \hline
        
        \multirow{3}{*}{DenseNet}   & Pure CNN          & 99.23  & 41.16 \\
        \cline{2-4}
        \multirow{3}{*}{}           &  \textbf{UNVP}    &  \textbf{99.42} & 41.98\\
        \cline{2-4}
        \multirow{3}{*}{}           &  \textbf{E-UNVP}  &  99.14 & \textbf{43.72} \\
        \hline
    \end{tabular}
\end{table}

\noindent \textbf{Backbone Deep Networks. } This section evaluates the robustness and the consistent improvements of UNVP and E-UNVP with common deep networks, including LeNet, AlexNet, VGG, ResNet, and DenseNet, as in Table \ref{tab:networks}. The proposed UNVP and E-UNVP consistently outperform the stand-alone classifier (Pure-CNN) using the same network configuration in all experiments. Particularly, it helps to improve \textbf{6\%}, \textbf{0.5\%}, \textbf{4\%}, \textbf{5\%}, \textbf{2\%} on MNIST-M using LeNet, AlexNet, VGG, ResNet and DenseNet respectively.

The proposed methods can be easily integrated with standard CNN deep networks. Therefore, it potentially can be applied to improve the performance in many existed CNN-based applications, e.g., detection and recognition, that are experimented in the next sections.

\begin{table}[!t]
    \small
    \centering
    \caption{Results $(\%)$ on three digit datasets. ADA and ours \textbf{do not require} target data in training. ADDA, DANN \textbf{require} training data from target domains in training.}
    \begin{tabular}{|c|c|c|c|c|}
         \hline
         \textbf{Methods} &  \textbf{MNIST} &  \textbf{SVHN} & \textbf{MNIST-M} \\
         \hline
         ADDA               & 99.29                                       & 32.20          & 63.39 \\
         DANN               & $-$                                         &  $-$           & 76.66 \\    
         \hline \hline
         Pure-CNN           & 99.06                                       & 31.96          & 55.90 \\
         ADA                & 99.17                                       & 37.87          & 60.02 \\
         
         \textbf{UNVP}      & 99.30 &  41.23 & 59.45 \\ 
         
         \textbf{E-UNVP}    & \textbf{99.42}  & \textbf{42.87} & \textbf{61.70} \\
         \hline
    \end{tabular}
    \label{tab:performance_on_digit_classification}
    \vspace{-5mm}
\end{table}

\noindent \subsection{Digit Recognition on Unseen Domains}
\label{sec:digit}

The proposed approaches have experimented in digit recognition on new unseen domains with two other digit databases, i.e., MNIST-M and SVHN (Fig. \ref{fig:char_dbs}). 
In this experiment, MNIST is the only database used to train the classifier. Then, two other datasets, i.e., MNIST-M and SVHN, are used as the new unseen domains to benchmark the performance.
The classifier is trained using 50,000 images of MNIST. In order to generalize an image phase, we use 10,000 images in this set to perturb and generalize new samples. All digit images are resized to $32 \times 32$. 
We benchmark the learned classifiers on MNIST and two other unseen digit datasets, i.e., SVHN and MNIST-M. The results using our approach are compared against the LeNet classifier (Pure-CNN), and the Adversarial Data Augmentation (ADA). We also show the recognition results on these datasets using the Domain Adaptation methods, including Adversarial Discriminative Domain Adaptation (ADDA), Domain-Adversarial Training of Neural Networks (DANN) \cite{pmlr-v37-ganin15}. It is noticed that Pure-CNN, ADA, and our approaches do not require the target domain data during training. Meanwhile, ADDA, DANN require the target domain data in the training steps.

Our generalization phase synthesizes images based on semantic space via the estimation of environment density. It helps our generated images to be more diverse than the synthesized images using the ADA method. The experimental results are shown in Table \ref{tab:performance_on_digit_classification}. 
The proposed methods consistently achieve state-of-the-art performance on these datasets. Notably, it helps to improve approximately \textbf{11\%} and \textbf{6\%} on SVHN and MNIST-M, respectively.

\begin{table}[!t]
\small
  \centering
  \caption{Results $(\%)$ on Extended Yale-B \cite{yale_b_dataset}, CMU-PIE \cite{pie_dataset} and CMU-MPIE \cite{multi_pie_dataset} databases. ADA and ours \textbf{do not require} target domain data during training while ADDA \textbf{does}.}
  \small
  \begin{tabular}{|c|c|c|c|c|c|c|}
    \hline
    \multirow{2}{*}{\textbf{Method}} &
      \multicolumn{2}{c|}{\textbf{E-Yale-B}} &
      \multicolumn{2}{c|}{\textbf{CMU-PIE}} &
      \multicolumn{2}{c|}{\textbf{CMU-MPIE}} \\ \cline{2-7}
                        & \textbf{N} & \textbf{D} & \textbf{N} & \textbf{D} & \textbf{N} & \textbf{D} \\
    \hline
    ADDA            & 99.17          & 75.28          & 96.09         & 70.33           & 99.93          & 97.71 \\
    \hline \hline
    Pure-CNN        & 98.50          & 51.39          & 95.59          & 62.18          & 99.93          & 94.74 \\
    ADA             & 99.00          & 53.08          & 96.49          & 62.69          & 99.92          & 96.08 \\
    \textbf{UNVP}   & 99.17          & 58.24          & 96.32          & 64.88          & 99.83          & \textbf{98.25} \\ 
    \textbf{E-UNVP} & \textbf{99.54} & \textbf{62.95} & \textbf{97.55} & \textbf{66.89} & \textbf{99.93} & 98.03 \\ 
    \hline
  \end{tabular}
    \label{tab:performance_on_face_dataset}
    \vspace{-5mm}
\end{table}

\subsection{Face Recognition on Unseen Domains}
\label{sec:face}

In this experiment, the proposed approaches are applied in unseen environment face recognition and compared against the other baseline methods, i.e., Pure-CNN, ADA, and ADDA, on three face recognition databases, including Extended Yale-B, CMU-PIE, and CMU-MPIE.
% (Fig. \ref{fig:face_dbs}). 
In each database, we select the face images with normal lighting as the source domain, i.e., Normal illumination (N), and the face images with dark lighting as the target domain, i.e., Dark illumination (D). 
Each database is randomly split into two sets: a training set (80\%) and a testing set (20\%). The experimental framework structures are similar to the one in digit recognition. All cropped face images are resized to $64 \times 64$ pixels. The experimental results in Table \ref{tab:performance_on_face_dataset} show that our proposed methods help to improve the recognition performance on new unseen domains where the lighting conditions are unknown. Particularly, it helps to improve approximately \textbf{11\%}, \textbf{4\%} and \textbf{3\%} in dark lighting conditions on Extended Yale-B, CMU-PIE and CMU-MPIE databases respectively.

\subsection{Pedestrian Recognition on Unseen Domains} \label{sec:ped}

This experiment aims to improve RGB-based pedestrian recognition on thermal images on the Thermal Dataset\footnote{\url{https://www.flir.com/oem/adas/adas-dataset-form/}}. %This dataset includes both thermal and RGB images. 
There are two datasets organized in this experiment: (1) RGB pedestrian and (2) Thermal pedestrian. 
The methods are trained only on the RGB pedestrian dataset and tested on the Thermal pedestrian dataset. 
In the training phase, we use $2,000$ images to generalize new images, and all images of two datasets are resized to $128 \times 128$ pixels.
%They also achieves higher accuracy than the baseline.
%As the results of Table \ref{tab:pedestrian_recognition} as well as experimental results on Digit Recognition and Face Recognition mentioned above, our proposals help our method outperform than Pure-CNN and ADA when testing on unseen domain.
% To further improve pedestrian detection, we apply our pedestrian recognition trained on RGB pedestrian dataset to the Deformable-CNNs detector \cite{dai17dcn, dai16rfcn} trained on COCO \cite{coco_dataset} dataset. After the image proposal phase, we crop proposed bounding boxes and feed into our pedestrian recognition framework.
The experimental results in Table \ref{tab:pedestrian_recognition} show that our proposed methods consistently help to improve the performance of the Pure-CNN in various common deep network structures, including LeNet, AlexNet, VGG, ResNet, and DenseNet. 
%Table \ref{tab:pedestrian_detection} shows our experimental results of pedestrian detection on Thermal Dataset. Because of the abstract pedestrian shape (body shape) is keep on the thermal image. Therefore, the vanilla Deformable-CNNs detector also have good results. Although our proposal helps to softly improve the results, the results prove that our method is promising.

\begin{table}[!t]
    \small
    \centering
    \caption{Results  $(\%)$ on RGB and Thermal pedestrian databases with various common deep network structures.}
    \label{tab:pedestrian_recognition}
    \begin{tabular}{|c|c|c|c|}
        \hline
        \textbf{Networks} & \textbf{Methods} & \textbf{RGB} & \textbf{Thermal} \\
        \hline
        \multirow{2}{*}{LeNet}      & Pure-CNN          & 95.45           &   79.72 \\
        \cline{2-4}
        \multirow{2}{*}{}           &  \textbf{E-UNVP}  & \textbf{97.25} & \textbf{90.29} \\
        \hline
        \multirow{2}{*}{AlexNet}    & Pure CNN          & 96.64           & 81.38 \\
        \cline{2-4}
        \multirow{2}{*}{}           &  \textbf{E-UNVP}  & \textbf{97.04}  & \textbf{82.98} \\
        \hline
        \multirow{2}{*}{VGG}        & Pure CNN          & 97.54           & 95.60  \\
        \cline{2-4}
        \multirow{2}{*}{}           &  \textbf{E-UNVP}  & \textbf{98.64}  &  \textbf{98.38}\\
        \hline
        \multirow{2}{*}{ResNet}     & Pure CNN          & 98.52           & 96.07 \\
        \cline{2-4}
        \multirow{2}{*}{}           &  \textbf{E-UNVP}  & \textbf{98.56}  & \textbf{98.35} \\
        \hline
        \multirow{2}{*}{DenseNet}   & Pure CNN          & 98.39           & 95.87 \\
        \cline{2-4}
        \multirow{2}{*}{}           &  \textbf{E-UNVP}  & \textbf{98.60}  & \textbf{96.14} \\
        \hline
    \end{tabular}
    \vspace{-5mm}
\end{table}

\section{Conclusions}
\label{sec:concl}

This paper has introduced the novel deep learning based domain generalization approach that generalizes well to different unseen domains. Only using training data from a source domain, we propose an iterative procedure that augments the dataset with samples from a fictitious target domain that is hard under the current model. It can be easily integrated with any other CNN based framework within an end-to-end network to improve the performance.
On digit recognition, the proposed method has been benchmarked on three popular digit recognition datasets and consistently showed the improvement. The method is also experimented in face recognition on three standard databases and outperforms the other state-of-the-art methods. In the problem of pedestrian recognition, we empirically observe that the proposed method learns models that improve performance across a priori unknown data distributions.

\section{Acknowledgement}
In this project, Dat T. Truong and Minh-Triet Tran are partially supported by Vingroup Innovation Foundation (VINIF) in project code VINIF.2019.DA19.




\bibliographystyle{IEEEtran}
\bibliography{references}

\begin{thebibliography}{10}
\providecommand{\url}[1]{#1}
\csname url@samestyle\endcsname
\providecommand{\newblock}{\relax}
\providecommand{\bibinfo}[2]{#2}
\providecommand{\BIBentrySTDinterwordspacing}{\spaceskip=0pt\relax}
\providecommand{\BIBentryALTinterwordstretchfactor}{4}
\providecommand{\BIBentryALTinterwordspacing}{\spaceskip=\fontdimen2\font plus
\BIBentryALTinterwordstretchfactor\fontdimen3\font minus
  \fontdimen4\font\relax}
\providecommand{\BIBforeignlanguage}[2]{{%
\expandafter\ifx\csname l@#1\endcsname\relax
\typeout{** WARNING: IEEEtran.bst: No hyphenation pattern has been}%
\typeout{** loaded for the language `#1'. Using the pattern for}%
\typeout{** the default language instead.}%
\else
\language=\csname l@#1\endcsname
\fi
#2}}
\providecommand{\BIBdecl}{\relax}
\BIBdecl

\bibitem{pmlr-v37-ganin15}
Y.~Ganin and V.~Lempitsky, ``Unsupervised domain adaptation by
  backpropagation,'' in \emph{ICML}, 2015.

\bibitem{adda_cvpr2017}
E.~Tzeng, J.~Hoffman, K.~Saenko, and T.~Darrell, ``Adversarial discriminative
  domain adaptation,'' July 2017.

\bibitem{triplet_loss}
F.~Schroff, D.~Kalenichenko, and J.~Philbin, ``Facenet: A unified embedding for
  face recognition and clustering,'' in \emph{CVPR}, June 2015.

\bibitem{range_loss}
X.~{Zhang}, Z.~{Fang}, Y.~{Wen}, Z.~{Li}, and Y.~{Qiao}, ``Range loss for deep
  face recognition with long-tailed training data,'' in \emph{ICCV}, 2017.

\bibitem{deep_resnet}
K.~{He}, X.~{Zhang}, S.~{Ren}, and J.~{Sun}, ``Deep residual learning for image
  recognition,'' in \emph{CVPR}, 2016.

\bibitem{DBLP:journals/corr/TzengHDS15}
E.~Tzeng, J.~Hoffman, T.~Darrell, and K.~Saenko, ``Simultaneous deep transfer
  across domains and tasks,'' \emph{CoRR}, 2015.

\bibitem{Sener:2016:LTR:3157096.3157333}
O.~Sener, H.~O. Song, A.~Saxena, and S.~Savarese, ``Learning transferrable
  representations for unsupervised domain adaptation,'' in \emph{NIPS}, 2016.

\bibitem{DBLP:journals/corr/TzengHZSD14}
E.~Tzeng, J.~Hoffman, N.~Zhang, K.~Saenko, and T.~Darrell, ``Deep domain
  confusion: Maximizing for domain invariance,'' \emph{CoRR}, 2014.

\bibitem{NIPS2016_6544}
M.-Y. Liu and O.~Tuzel, ``Coupled generative adversarial networks,'' in
  \emph{Advances in Neural Information Processing Systems 29}, 2016.

\bibitem{feature_transfer_learning}
X.~Yin, X.~Yu, K.~Sohn, X.~Liu, and M.~Chandraker, ``Feature transfer learning
  for deep face recognition with long-tail data,'' \emph{CoRR}, 2018.

\bibitem{Reynolds00speakerverification}
D.~A. Reynolds, T.~F. Quatieri, and R.~B. Dunn, ``Speaker verification using
  adapted gaussian mixture models,'' in \emph{Digital Signal Processing}, 2000.

\bibitem{Murez_2018_CVPR}
Z.~Murez, S.~Kolouri, D.~Kriegman, R.~Ramamoorthi, and K.~Kim, ``Image to image
  translation for domain adaptation,'' in \emph{CVPR}, June 2018.

\bibitem{TzengHSD17}
E.~Tzeng, J.~Hoffman, K.~Saenko, and T.~Darrell, ``Adversarial discriminative
  domain adaptation,'' \emph{CVPR}, 2017.

\bibitem{saito2018maximum}
K.~Saito, K.~Watanabe, Y.~Ushiku, and T.~Harada, ``Maximum classifier
  discrepancy for unsupervised domain adaptation,'' in \emph{CVPR}, 2018.

\bibitem{shankar2018generalizing}
S.~Shankar, V.~Piratla, S.~Chakrabarti, S.~Chaudhuri, P.~Jyothi, and
  S.~Sarawagi, ``Generalizing across domains via cross-gradient training,''
  2018.

\bibitem{generalize-unseen-domain}
R.~Volpi, H.~Namkoong, O.~Sener, J.~C. Duchi, V.~Murino, and S.~Savarese,
  ``Generalizing to unseen domains via adversarial data augmentation,''
  \emph{NIPS}, 2018.

\bibitem{domain_generalization_Ghifary_2015_ICCV}
M.~Ghifary, W.~Bastiaan~Kleijn, M.~Zhang, and D.~Balduzzi, ``Domain
  generalization for object recognition with multi-task autoencoders,'' in
  \emph{ICCV}, 2015.

\bibitem{domain_generalization_Li_2018_CVPR}
H.~Li, S.~Jialin~Pan, S.~Wang, and A.~C. Kot, ``Domain generalization with
  adversarial feature learning,'' in \emph{CVPR}, 2018.

\bibitem{domain_generalization_Muandet_ICML_2018}
K.~Muandet, D.~Balduzzi, and B.~Schölkopf, ``Domain generalization via
  invariant feature representation,'' in \emph{ICML}, 2013.

\bibitem{domain_generalization_Li_2018_ECCV}
Y.~Li, X.~Tian, M.~Gong, Y.~Liu, T.~Liu, K.~Zhang, and D.~Tao, ``Deep domain
  generalization via conditional invariant adversarial networks,'' in
  \emph{ECCV}, 2018.

\bibitem{lenet_ref}
Y.~Lecun, L.~Bottou, Y.~Bengio, and P.~Haffner, ``{Gradient-based learning
  applied to document recognition},'' \emph{Proceedings of the IEEE}, 1998.

\bibitem{deep_alexnet}
A.~Krizhevsky, I.~Sutskever, and G.~E. Hinton, ``Imagenet classification with
  deep convolutional neural networks,'' in \emph{NIPS}, 2012.

\bibitem{deep_vgg}
K.~Simonyan and A.~Zisserman, ``Very deep convolutional networks for
  large-scale image recognition,'' in \emph{ICLR}, 2015.

\bibitem{deep_densenet}
G.~Huang, Z.~Liu, L.~van~der Maaten, and K.~Q. Weinberger, ``Densely connected
  convolutional networks,'' in \emph{CVPR}, 2017.

\bibitem{glow_ref}
D.~P. Kingma and P.~Dhariwal, ``Glow: Generative flow with invertible 1x1
  convolutions,'' in \emph{NIPS}, 2018.

\bibitem{svhn_dataset}
Y.~Netzer, T.~Wang, A.~Coates, A.~Bissacco, B.~Wu, and A.~Y. Ng, ``Reading
  digits in natural images with unsupervised feature learning,'' in
  \emph{NIPSW}, 2011.

\bibitem{yale_b_dataset}
A.~Georghiades, P.~Belhumeur, and D.~Kriegman, ``From few to many: Illumination
  cone models for face recognition under variable lighting and pose,''
  \emph{TPAMI}, 2001.

\bibitem{pie_dataset}
T.~Sim, S.~Baker, and M.~Bsat, ``The cmu pose, illumination, and expression
  (pie) database,'' in \emph{FG}, 2002.

\bibitem{multi_pie_dataset}
R.~Gross, I.~Matthews, J.~Cohn, T.~Kanade, and S.~Baker, ``Multi-pie,''
  \emph{IVC}, 2010.

\end{thebibliography}



% \end{thebibliography}

\end{document}